\documentclass[sigconf]{acmart}

\usepackage{bm}
\usepackage{color, soul}
\usepackage{framed}
\usepackage{color}
\definecolor{shadecolor}{rgb}{0.92,0.92,0.92}

\usepackage{balance}

\def\BibTeX{{\rm B\kern-.05em{\sc i\kern-.025em b}\kern-.08emT\kern-.1667em\lower.7ex\hbox{E}\kern-.125emX}}
    
\copyrightyear{2019}
\acmYear{2019}
\acmConference[CIKM '19]{The 28th ACM International Conference on Information and
Knowledge Management}{November 3--7, 2019}{Beijing, China}
\acmBooktitle{The 28th ACM International Conference on Information and Knowledge
Management (CIKM '19), November 3--7, 2019, Beijing, China}
\acmPrice{15.00}
\acmDOI{10.1145/3357384.3358165}
\acmISBN{978-1-4503-6976-3/19/11}

\setcopyright{acmcopyright}
\begin{document}
\sethlcolor{yellow}
\newcommand{\tabincell}[2]{\begin{tabular}{@{}#1@{}}#2\end{tabular}}
\fancyhead{}

\title{Incorporating Relation Knowledge into Commonsense Reading Comprehension with Multi-task Learning}


\author{Jiangnan Xia, Chen Wu, Ming Yan}
\orcid{1234-5678-9012}
\affiliation{%
  \institution{Alibaba DAMO Academy}
  \city{Hangzhou}
  \state{China}
}
\email{{jiangnan.xjn, wuchen.wc, ym119608}@alibaba-inc.com}





%
\renewcommand{\shortauthors}{Xia, et al.}

%
\begin{abstract}
This paper focuses on how to take advantage of external relational knowledge to improve machine reading comprehension (MRC) with multi-task learning. Most of the traditional methods in MRC assume that the knowledge used to get the correct answer generally exists in the given documents. However, in real-world task, part of knowledge may not be mentioned and machines should be equipped with the ability to leverage external knowledge. 
In this paper, we integrate relational knowledge into MRC model for commonsense reasoning. Specifically, based on a pre-trained language model (LM), we design two auxiliary relation-aware tasks 
to predict if there exists any commonsense relation and what is the relation type between two words, 
in order to better model the interactions between document and candidate answer option. We conduct experiments on two multi-choice benchmark datasets: the SemEval-2018 Task 11 and the Cloze Story Test. The experimental results demonstrate the effectiveness of the proposed method, 
which achieves superior performance compared with the comparable baselines on both datasets. 
\end{abstract}

%
%



%
\maketitle

\section{Introduction}
Machine reading comprehension (MRC) enables machines with the ability to answer questions with given corresponding documents. 
Recent years have witnessed the bloom of various well-designed MRC models~\cite{wang2017gated, yu2018qanet,wang2018multi}, which achieve promising performance when provided with adequate manually labeled instances~\cite{rajpurkar2016squad}. However, these models generally assume that the knowledge required to answer the questions has already existed in the documents, which does not hold at some time. How to leverage the commonsense knowledge for better reading comprehension remains largely unexplored. 

Recently, some preliminary studies have begun to incorporate certain side information (e.g., triplets from external knowledge base) into the model design of various NLP tasks, such as question answering~\cite{bauer2018commonsense} and conversation generation~\cite{zhou2018commonsense}. Generally, there are two lines of this work. The first 
line focuses on designing task-specific model structures~\cite{yang2017leveraging,bauer2018commonsense}, which exploit the retrieved concepts from external knowledge base for enhancing the representation. Recently, the other line has studied to pre-train a language model over large corpus to learn the inherent word-level knowledge in an unsupervised way~\cite{radford2018improving, devlin2018bert}, which achieves very promising performance. 

The first line of work is usually carefully designed for the target task, which is not widely applicable. The second line can only learn the co-occurrence of words or entities in the context, while it may not be that robust for some complex scenarios such as \textit{reasoning} task. For example, to answer the question \textit{"Was the light bulb still hot?"} 
when the document is given as \textit{"I went into my bedroom and flipped the light switch. Oh, I see that the ceiling lamp is not turning on..."}, 
machines should have the commonsense knowledge that 
\textit{"the bulb is not hot when turning off"} to correctly answer the question. The explicit relation information can act as a bridge to connect the scattered context, which may not be easily captured. Therefore, the aim of this paper is to take the advantage of both the pre-trained language model and the explicit relation knowledge from the external knowledge base for commonsense reading comprehension. 

Specifically, we first extract the triplets from the popular ConceptNet knowledge base~\cite{speer2017conceptnet} 
and design two auxiliary relation-aware tasks to predict if there exists any relation and what is the relation type between two concepts.
To make the model be aware of the commonsense relations between concepts, we propose a multi-task learning framework to jointly learn the prediction of the target MRC task and the two relation-aware tasks in a unified model. 
We conduct experiments on two multi-choice commonsense reading comprehension datasets: 
Story Cloze Test \cite{mostafazadeh2017lsdsem} and 
SemEval-2018 Task 11 \cite{ostermann2018semeval}. 
Experimental results demonstrate the effectiveness of our method, 
which achieves superior performance compared with the comparable baselines on both datasets.

\section{Related Work}
Previous studies mainly focused on developing effective model structures to improve the reading 
ability of the systems \cite{wang2017gated, yu2018qanet}, 
which have achieved promising performance. 
However, the success on these tasks is not adequate considering the model's ability of commonsense reasoning. 
Recently, a number of efforts have been invested in developing datasets for commonsense reading comprehension such as Story Cloze Test and SemEval-2018 Task 11~\cite{ostermann2018semeval, mostafazadeh2017lsdsem}. 
In these datasets, part of required knowledge may not be mentioned in the document and machines should be equipped with commonsense knowledge to make correct prediction. 
There exists increasing interest in incorporating commonsense knowledge into commonsense reading comprehension tasks. 
Most of previous studies focused on developing 
special model structures to introduce external knowledge 
into neural network models \cite{yang2017leveraging,bauer2018commonsense}, 
which have achieved promising results. 
For example, Yang and Mitchell \cite{yang2017leveraging} use
concepts from WordNet and weighted average vectors of the retrieved concepts to calculate a new LSTM state. 
These methods relied on task-specific model structures which are 
difficult to adapt to other tasks. 
Pre-trained language model such as BERT and GPT\cite{radford2018improving, devlin2018bert} is also used as a kind of commonsense knowledge source. However, the LM method mainly captures the co-occurrence of words and phrases and cannot address some more complex problems which may require the reasoning ability.

Unlike previous work, we incorporate external knowledge by jointly training MRC 
model with two auxiliary tasks which are relevant to commonsense knowledge.
The model can learn to fill in the knowledge gap without changing the original model structure. 


\section{Knowledge-enriched MRC model}
\subsection{Task Definition}
Here we formally define the task of multi-choice commonsense  reading comprehension. Given a reference document $\bm{D}$ (a question $\bm{q}$ if possible),  
a set of $N$ answer options $\{\bm{O}_{1}, \bm{O}_{2},..., \bm{O}_{N}\}$  and an external knowledge base $\textbf{F}=\{f_1, \cdots, f_M\}$, the goal is to choose the correct answer option according to their probabilities $\{p_{1}, p_{2},..., p_{N}\}$ given by MRC model, where $N$ is the total number of options. 

In this paper, we use ConceptNet knowledge base~\cite{speer2017conceptnet},   
which is a large semantic network of commonsense knowledge with a total of about 630k facts. 
Each fact $f_i$ is represented as a triplet $f_{i}=(subject, relation, object)$, 
where $subject$ and $object$ can be a word or phrase and $relation$ is a relation type. An example is:
 $([Car]_{subj}, [UsedFor]_{rel}, [Driving]_{obj})$


\subsection{Overall Framework}
The proposed method can be roughly divided into three parts: 
a pre-trained LM encoder, a task-specific prediction layer for multi-choice MRC and two relation-aware auxiliary tasks. 
The overall framework is shown in Figure 1. 

The pre-trained LM encoder acts as the foundation of the model, which is used to capture the relationship between question, 
document and answer options. Here we utilize 
BERT \cite{devlin2018bert} as the pre-trained encoder for its superior performance in a range of natural language understanding tasks. Specially, we concatenate the given document, question (as sentence A) and each option (as sentence B) with special delimiters as one segment, which is then fed into BERT encoder. The input sequence is packed as ``[CLS]D(Q)[SEP]O[SEP]''~\footnote{\small{We directly concatenate the question after the document to form the BERT input if a question is given.}}, where [CLS] and [SEP] are the special delimiters. After BERT encoder, we obtain the contextualized word representation $\bm{h}^{L}_{i} \in \mathbb{R}^H$ for the $i$-th input token from the final layer of BERT. $H$ is the dimension of hidden state.

Next, on top of BERT encoder, we add a task-specific output layer and view the multi-choice MRC as a multi-class classification task. Specifically, we apply a linear head layer plus a softmax layer on the final contextualized word representation of [CLS] token $\bm{h}^{L}_{0}$. We minimize the Negative Log Likelihood (NLL) loss with respect to the correct class label, as:  
\begin{equation}\label{equ:1}
  \begin{split}
    L^{AP}(\hat{\bm{O}}|\bm{D}) = -log\frac{exp(\bm{v}^T\bm{\hat{h}}_0)}{\Sigma^N_{k=1}exp(\bm{v}^{T}\bm{h}^{k}_0)}
  \end{split}
\end{equation}
where $\hat{\bm{h_0}}$ is the final hidden state of the correct option $\hat{\bm{O}}$, $N$ is the number of options 
and $\bm{v}^T \in \mathbb{R}^H$ is a learnable vector. 

 \begin{figure}
  \centering
  \includegraphics[width=0.5\textwidth]{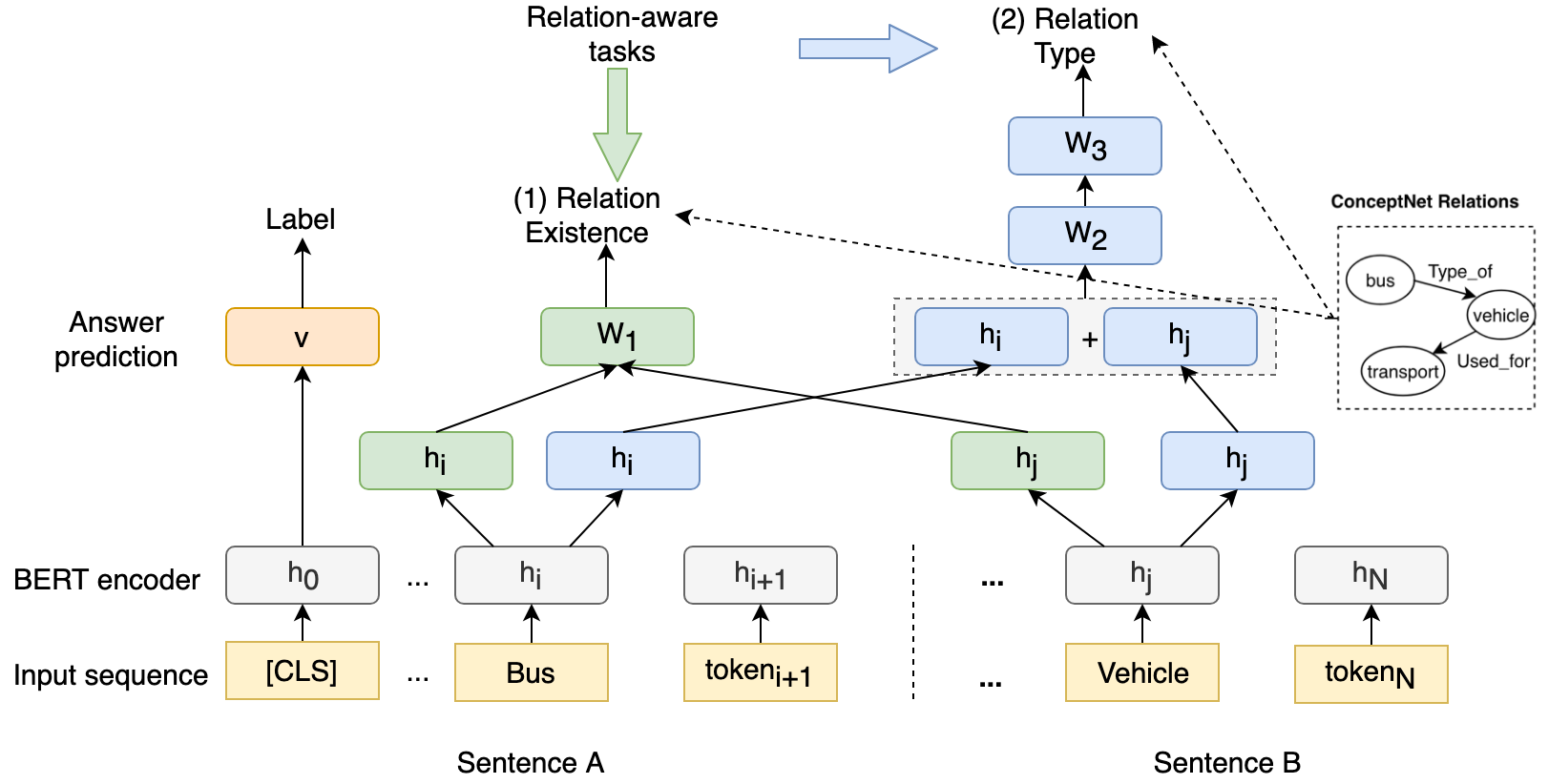}\vspace{-2mm}
  \caption{MRC model with two relation-aware tasks}\vspace{-2mm}
  \label{fig:cascade}
\end{figure}

Finally, to make the model be aware of certain implicit commonsense relations between concepts, we further introduce two auxiliary relation-aware prediction tasks for joint learning. Since it may be  difficult to directly predict the actual relation type between two concepts without adequate training data, we split the relation prediction problem into two related tasks: i.e., \textit{relation-existence} task and \textit{relation-type} task. In \textit{relation-existence}, we basically add an auxiliary task to predict if there exists any relation between two concepts, which is a relatively easy task. Then, we take one step further to decide what is the right type of the relation in \textit{relation-existence}. The basic premise is that by guiding the MRC model training with extra relation information, the proposed model can be equipped with the ability to capture some underlying commonsense relationships. The two auxiliary tasks are jointly trained with the multi-choice answer prediction task. In the following, we will describe the two auxiliary tasks in detail.


\subsection{Incorporating Relation Knowledge}


\textbf{Task 1} is the \textit{relation-existence} task. 
Following \cite{devlin2018bert}, 
we first convert the concept to a set of BPE tokens tokens A and tokens B, 
with beginning index $i$ and $j$ in the input sequence respectively. 
The probability of whether there is a relation in each pair (tokens A, tokens B) is computed as: 
\begin{equation}\label{equ:2}
  \begin{split}
    p^{RE}_{ij}={\rm sigmoid}(\bm{h}^{T}_{i} \bm{W}_{1} \bm{h}_{j})
  \end{split}
\end{equation} 
where $\bm{W}_{1} \in \mathbb{R}^{H \times H}$ is a trainable matrix. 

We define the pair (tokens A, tokens B) that has relation in ConceptNet 
as a positive example and others as negative examples. 
We down-sample the negative examples and keep ratio of positive \textit{vs} negative is $1:\gamma$. 
We define the \textit{relation-existence} loss as the average binary cross-entropy (BCE) loss: 
\begin{equation}\label{equ:3}
  \begin{split}
    &{\mathcal{L}^{RE}}= \frac{1}{|A|}\frac{1}{|B|}\sum^{|A|}_{i=1}\sum^{|B|}_{j=1}{\rm BCE}(y^{RE}_{ij}, p^{RE}_{ij})
  \end{split}
\end{equation}
where $|A|$, $|B|$ are the number of sampled concepts in sentence A and sentence B respectively, 
$y^{RE}$ is the label of whether there is a relation between concepts. 

\textbf{Task 2} is the \textit{relation-type} task.
We predict the relation type between tokens A and tokens B. 
The \textit{relation-type} probabilities are computed as: 
\begin{equation}\label{equ:4}
  \begin{split}
    \bm{p}^{RT}_{ij}={\rm softmax}(\bm{W}_{3}{\rm ReLU}(\bm{W}_{2}[\bm{h}_i;\bm{h}_j])) \\
  \end{split}
\end{equation} 
where $\bm{W}_{2} \in \mathbb{R}^{H \times 2H}$ and $\bm{W}_{3} \in \mathbb{R}^{R \times H}$
are new trainable matrices, 
$R$ is the number of selected relation types \footnote{We don't use the entire set of 
relation types because not all the types are 
defined clearly. 
We choose 34 kind of relations 
except "RelatedTo", "ExternalURL" and "dbpedia".}.


The \textit{relation-type} loss is computed as: 
\begin{equation}\label{equ:5}
  \begin{split}
    &\mathcal{L}^{RT} = -\frac{1}{|S|}\sum^{|A|}_{i=1} \sum^{|B|}_{j=1} s_{ij} {\rm log} [ {{\bm{p}^{RT}_{ij}}} ] _k
  \end{split}
\end{equation} 
We define $s_{ij}$ as the label whether there is a relation from sentence A to sentence B. 
$k$ is the index of ground-truth relation in ConceptNet, 
$|S|$ is the number of relations among the tokens in two sentences. 

As the three tasks share the same BERT architecture with only different linear head layers, 
we propose to train them together. 
The joint objective function is formulated as follows:
\begin{equation}\label{equ:6}
  \begin{split}
    \mathcal{L} = \mathcal{L}^{AP} + \frac{1}{N} \sum^{N}_{l=1} (\lambda_{1} \mathcal{L}^{RE}_{l} + \lambda_{2} \mathcal{L}^{RT}_{l})
  \end{split}
\end{equation} 
where $\lambda_{1}$ and $\lambda_{2}$ are two 
hyper-parameters that control the weight of the tasks, 
$N$ is number of options. 

\section{Experiments}
\subsection{Dataset}
We conduct experiments on two commonsense reading comprehension tasks: 
SemEval-2018 shared task11 \cite{ostermann2018semeval} and Story Cloze Test \cite{mostafazadeh2017lsdsem}. 
The statistics of the datasets are shown in Table 1 \footnote{Story Cloze Test consists 98,161 unlabeled examples 1,871 labeled examples. Following \cite{radford2018improving}, 
we divide the labeled examples to a new training set and development set with 
1,433 and 347 examples respectively.}. 

\begin{table}
  \caption{Number of examples in datasets.}
  \label{tab:table1}
  \begin{tabular}{cccc}
  \toprule
    \textbf{Dataset}&Train &Dev &Test\\
  \midrule
    SemEval-2018 Task11 & 9,731 & 1,411 & 2,797 \\
    Story Cloze Test & 1,433 & 347 & 1,871\\
  \bottomrule
\end{tabular}
\end{table}

\begin{table}
  \caption{Performance on SemEval-2018 Task 11.}
  \centering
  \label{tab:table2}
  \begin{tabular}{lc}
    \toprule
    Model & ACC.(Test) \% \\
    \midrule
    NN-T \cite{merkhofer2018mitre}  & 80.23 \\
    HMA \cite{chen2018hfl} & 80.94 \\
    TriAN \cite{wang2018yuanfudao}  & 81.94 \\
    TriAN + $f^{dir}_{cs}$ + $f^{ins}_{cs}$ \cite{zhong2018improving} & 81.80 \\
    \hline
    TriAN + relation-aware tasks & 82.84 \\
    BERT(base) & 87.53 \\
    \textbf{BERT(base) + relation-aware tasks} & \textbf{88.23} \\
    \bottomrule
\end{tabular}
\end{table}

\begin{table}
  \caption{Performance on Cloze Story Test.} 
  \label{tab:table3}
  \begin{tabular}{lc}
    \toprule
    Model & ACC.(Test) \% \\
    \midrule
    Style+RNNLM \cite{schwartz2017story} & 75.2 \\
    HCM \cite{chaturvedi2017story} & 77.6 \\
    GPT \cite{radford2018improving} & 86.5 \\
    \hline
    BERT(base) & 86.7 \\
    \textbf{BERT(base) + relation-aware tasks} & \textbf{87.4} \\
    \bottomrule
\end{tabular}
\end{table}

\begin{table}
  \caption{Performance with different tasks}
  \label{tab:table4}
  \begin{tabular}{lcc}
    \toprule
    Model&ACC.(Dev)\% &$\Delta$\\
    \midrule
    \textbf{Basic Model} & 87.51 & - \\
    \hline
    + $\mathcal{L}_{RE}$ & 88.02 & +0.51 \\
    + $\mathcal{L}_{RT}$ & 87.87 & +0.36 \\
    + $\mathcal{L}_{RE}$ + $\mathcal{L}_{RT}$ & 88.25 & +0.74 \\
    + $\mathcal{L}_{RT}$ + "No Relation" & 87.78 & +0.41 \\
  \bottomrule
\end{tabular}

\end{table}

\begin{table*}
  \caption{Examples that require commonsense relations between concepts}
  \label{tab:table5}
  \begin{tabular}{|p{7.5cm}|p{2cm}|p{4cm}|p{3cm}|}
    \hline
    \textbf{Document} & \textbf{Question} & \textbf{Options} & \textbf{Commonsense Facts} \\
    \hline
    [SemEval] ... We organized it from the fruit to the dairy in an organized order.
  We put the shopping list on our fridge so that we wouldn't forget it the next day when we went to go buy the food. & What did they  \textbf{write} the list on? & (A) \textbf{Paper.}  (B) Fridge.  Correct: A &  [\textbf{paper}, RelatedTo, \textbf{write}] \sethlcolor{yellow} \\
    \hline
    [Cloze Story Test] My kitchen had too much trash in it. I \textbf{cleaned it up} and put it into bags. I took the bags outside of my house. I then carried the bags down my driveway to the trash can. & - & 
    (A) I missed the trash in the kitchen.  (B) I was glad to \textbf{get rid of} the trash.  Correct: B & [get rid of, RelatedTo, clean up] \\
    \hline
    [SemEval] ... I settled on Earl Gray, which is a black tea flavored with bergamot orange. I filled the \textbf{kettle} with water and placed it on the stove, turning on the burner...
    & Why did they \textbf{use} a kettle? & 
    (A) To drink from. (B) To \textbf{boil water.} Correct: B
    & [kettle, \textbf{UsedFor}, boil water]; [kettle, RelatedTo, drinking] \\
    \hline
\end{tabular}
\end{table*}

\subsection{Implementation Details}
We use the uncased BERT(base) \cite{devlin2018bert} as pre-trained language model. 
We set the batch size to 24, learning rate to 2e-5. 
The maximum sequence length is 384 for SemEval-2018 Task 11 and 512 for Story Cloze Test.  
We fine-tune for 3 epochs on each dataset. 
The task-specific hyper-parameters $\lambda_{1}$ and $\lambda_{2}$ are set to 0.5 
and the ratio $\gamma$ is set to 4.0.
\subsection{Experimental Results and Analysis}
\textbf{Model Comparison} The performances of our model on two datasets are shown in Table 2 and Table 3. 
We compare our single model with other existing systems (single model). 
We also adopt relation-aware tasks on the co-attention layer of the TriAN model \cite{wang2018yuanfudao}. 
From the result, we can observe that: 
(1) Our method achieves better performance 
on both datasets compared with previous methods and Bert(base) model. 
(2) By adopt relation-aware tasks on the attention layer of the TriAN model \cite{wang2018yuanfudao} on SemEval, the model performance can also be improved. 
The results show that the relation-aware tasks can help to better align sentences due to knowledge gap. 
\\
\textbf{Effectiveness of Relation-aware Tasks}  To get better insight into our model, 
we analyze the benefit brought by using 
relation-aware tasks 
on the development set of SemEval-2018 Task 11. 
The performance of jointly training the basic answer prediction model with different tasks is shown in table 4. 
From the result we can see that by incorporating the auxiliary 
\textit{relation-existence} task ($\mathcal{L}_{RE}$) or \textit{relation-type} 
task ($\mathcal{L}_{RT}$) in the joint learning framework, the performance can always improved. 
The result shows the advantage of incorporating auxiliary tasks. 
Besides, the performance gain by adding \textit{relation-existence} task is larger, 
which shows \textit{relation-existence} task can incorporate more knowledge into the model. 
We also attempt to merge two relation-aware tasks into one task 
by simply taking "No-Relation" as a special type of relation. 
The model performance is just slightly higher than using \textit{relation-type} task 
and lower than using two tasks separately. 
The result is due to the number of "No-Relation" labels is much more than other relation types, 
which makes the task hard to train. 
\\
\textbf{Analysis}
Table 5 shows the 
examples that are incorrectly 
predicted by BERT(base), while correctly solved by incorporating relation-aware tasks.
The first two examples can benefit from \textit{relation-existence} knowledge. 
From the first example we can see that the retrieved relation between concepts from ConceptNet 
provide useful evidence to connect the question to the correct option (A). 
The second example is from Cloze Story Test dataset, we can see that the retrieved relation 
is also helpful in making correct prediction. 
The third example from SemEval requires \textit{relation-type} knowledge. 
but the relation type $(kettle, UsedFor, boil water)$ in option (A), 
is more relevant to the question, which shows that relation type knowledge 
can be used as side information to do the prediction.

\section{Conclusion}
In this paper, we aim to enrich the neural model with external knowledge to improve commonsense reading comprehension. 
We use two auxiliary relation-aware tasks to incorporate ConceptNet knowledge 
into the MRC model. 
Experimental results demonstrate the effectiveness of our method 
which achieves improvements 
compared with the pre-trained language model baselines on both datasets. 



\bibliographystyle{ACM-Reference-Format}
\bibliography{cikm}

\end{document}